\documentclass[10pt,twocolumn,letterpaper]{article}

\usepackage{cvpr}              

\usepackage{amssymb}
 \usepackage{amsfonts}
\usepackage{amsmath}
\usepackage{booktabs}
\usepackage{multirow}
\usepackage{multicol}
\usepackage{sidecap}
\usepackage{makecell}
\usepackage{diagbox}
\usepackage[table]{xcolor}
\usepackage{subcaption}
\usepackage{mathtools}
\usepackage{pdflscape}
\usepackage{pgfplots}
\usepackage{dsfont}
\usepackage{adjustbox}
\usepackage{pifont}
\definecolor{cvprblue}{rgb}{0.21,0.49,0.74}
\usepackage[pagebackref,breaklinks,colorlinks,allcolors=cvprblue]{hyperref}

\title{From Spots to Pixels: Dense Spatial Gene Expression Prediction from Histology Images}

\author{Ruikun Zhang$^{1}$, \quad Yan Yang$^{2}$, \quad and \quad Liyuan Pan$^{1}$\\
$^{1}$Beijing Institute of Technology \quad $^{2}$ The Australian National University \\
{\tt\small \{ruikun.zhang, liyuan.pan\}@bit.edu.cn, \quad \{yan.yang\}@anu.edu.au}
}

\begin{document}
\maketitle

\begin{abstract}
Spatial transcriptomics (ST) measures gene expression at fine-grained spatial resolution, offering insights into tissue molecular landscapes. Previous methods for spatial gene expression prediction typically crop spots of interest from histopathology slide images, and train models to map each spot to a corresponding gene expression profile. 
However, these methods inherently lose the spatial resolution in gene expression:
1) each spot often contains multiple cells with distinct gene expression profiles;
2) spots are typically defined at fixed spatial resolutions, limiting the ability to predict gene expression at varying scales. 
To address these limitations, this paper presents PixNet, a dense prediction network capable of predicting spatially resolved gene expression across spots of varying sizes and scales directly from histopathology slide images. Different from previous methods that map individual spots to gene expression values, we generate a spatially dense continuous gene expression map from the histopathology slide image, and aggregate values within spots of interest to predict the gene expression. Our PixNet outperforms state-of-the-art methods on four common ST datasets in multiple spatial scales. The source code will be publicly available.
\end{abstract}

\section{Introduction}
\label{sec:intro}

Spatial transcriptomics (ST) enables spatially resolved gene expression profiling and offers significant clinical benefits, but the experimental process remains costly and technically demanding. Therefore, numerous studies have explored predicting spatial gene expression directly from more accessible and cost-effective histopathology slide images \cite{stnet,histogene,egn,eggn2023,bleep,triplex,sgn}. To retain spatial information, researchers have formulated the prediction task as a regression problem that maps spots of interest from histopathology slides to their corresponding gene expression.

\begin{figure}[t]
    \begin{tikzpicture}
    \tikzset{font=\small}
    \node[anchor=south west,inner sep=0] (image) at (0,0) {\includegraphics[width=\linewidth]{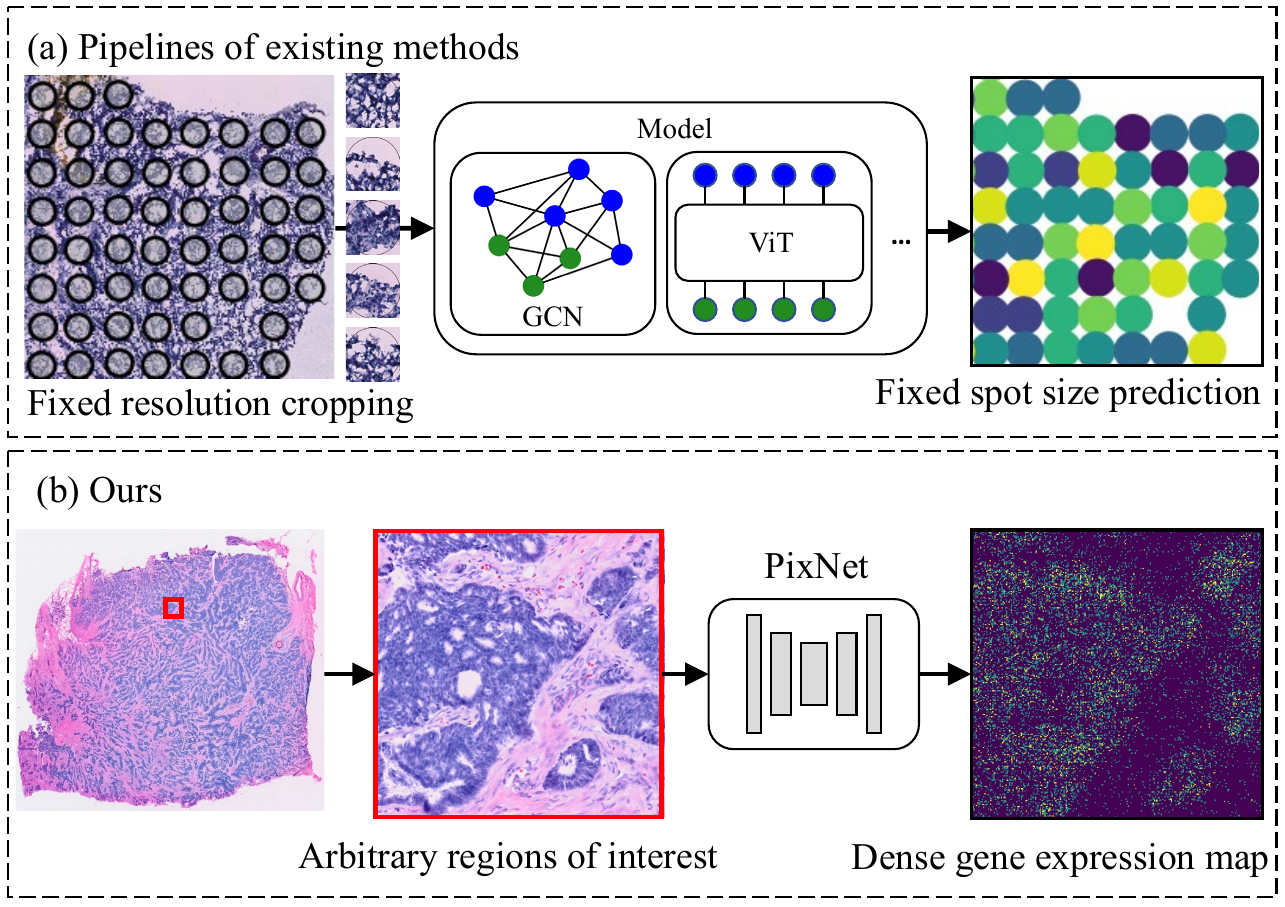}};
        \begin{scope}[x={(image.south east)},y={(image.north west)}]
        \end{scope}
    \end{tikzpicture}
    \caption{Overview of fields. (a) Existing approaches treat spatial gene expression prediction as a regression problem, training various networks on fixed crops from a slide image. (b) Our method formulates it as a dense prediction task, generating a gene expression map and aggregating values within spots of interest. 
    } 
    \label{fig:overview}
\end{figure}

Formally, previous works address this task by cropping spots with measured gene expression from available datasets and training networks on the resulting paired data for prediction. Various network architectures have been explored, including fine-tuning models pretrained on large-scale datasets \cite{egn,eggn2023}, as well as designing multi-scale \cite{triplex} and graph-based models \cite{sgn} to exploit the spatial context of histopathology slides and enhance prediction performance. An overview of these approaches is illustrated in Fig.~\ref{fig:overview} (a). While these approaches have demonstrated promising results, they fall short of achieving truly spatially resolved gene expression prediction. Specifically, the cropped spots typically exceed 100 $\mu$m in size to capture sufficient visual and spatial information for model input. As a result, these models aggregate features from multiple cells within each spot and associate a fixed crop with a single expression profile, leading to a loss of spatial resolution and, consequently, a diminished capability to predict spatially resolved gene expression.

Moreover, previous methods are typically designed for spots with a fixed spatial scale. However, real-world applications often require gene expression prediction across varying spatial resolutions and spot sizes, which limits the adaptability of existing methods. For example, a model trained on 100 $\mu$m spots struggles to generalize to 2 $\mu$m spots, as it depends on larger areas to capture sufficient spatial context for accurate prediction. Furthermore, 2 $\mu$m spots offer approximately single-cell resolution—the finest level of detail typically sought in spatial transcriptomics \cite{st_limitation_scale_0}. In practice, however, this corresponds to just $\sim 20$ pixels in the image space. This resolution constraint further restricts the applicability of existing methods in emerging ST technologies, such as Visium HD, which provides spatial gene expression at 2 $\mu$m spots. The reliance on large visual fields makes it difficult for these methods to adapt to such high-resolution advances.

To address these challenges, we reformulate spatial gene expression prediction from slide images as a dense prediction task. We propose PixNet, a model that maps the entire slide image to a dense gene expression map. For arbitrary regions of interest, gene expression is predicted by aggregating values from the corresponding area of the dense expression map. Fig.~\ref{fig:overview} (b) provides a visual summary of the proposed approach. We leverage the inherent multi-scale nature of slide images \cite{triplex} to generate the expression map. To capture features at multiple spatial scales, we extract a pyramidal feature map and progressively decode it into a high-resolution gene expression map, enabling dense spatial predictions. During training, we employ a sparse loss module that provides supervision only at spots with measured gene expression, effectively handling the inherent sparsity of spatial transcriptomics data. Our generalized approach enables accurate prediction of spatial gene expression across multiple scales and resolutions. We evaluate our method on four common ST datasets and demonstrate its effectiveness in predicting gene expression at scales and spot sizes different from those seen during training, outperforming previous methods. For example, when trained on 100 $\mu$m spots, PixNet generalizes to single-cell resolution (2 $\mu$m), achieving a Pearson correlation coefficient of 0.198—57.1\% higher than the previous best approach.

Our key contributions can be summarized as follows:
\begin{itemize}
    \item We reformulate spatial gene expression prediction as a dense prediction task. 

    \item We introduce a sparse supervision strategy designed to accommodate the limited coverage of spatial transcriptomics data. 

    \item We demonstrate state-of-the-art performance on four widely used ST datasets in multiple spatial scales.
\end{itemize}

\section{Related work}

In this section, we first review deep learning-based approaches for predicting spatial gene expression from whole slide images (WSIs). We then summarize relevant research on dense prediction networks in medical imaging that has informed the architectural design of our PixNet.
\paragraph{Spatial Gene Expression Prediction from WSIs.}

Due to the high cost and low throughput of spatial transcriptomics (ST), recent studies have leveraged deep learning to predict spatial gene expression from whole slide images (WSIs). ST-Net \cite{stnet} fine-tune a DenseNet-121 \cite{densenet} pretrained on ImageNet \cite{imagenet}, using histology–gene expression pairs for supervised learning. HisToGene \cite{histogene} introduces spatial context via a Vision Transformer, enabling context-aware feature learning. Hist2ST \cite{His2ST} further enhance performance by extracting patch-level features and modeling spatial dependencies with GraphSAGE \cite{graphSAGE}.


EGN \cite{egn,eggn2023} employs an exemplar-based framework that dynamically selects the most similar samples for each target spot, enhancing prediction accuracy. BLEEP \cite{bleep}, inspired by CLIP \cite{clip}, uses gene expression embeddings to guide the pretraining of a ResNet-50 \cite{resnet} image encoder, learning biologically meaningful visual features. TRIPLEX \cite{triplex} adopts a multi-scale strategy with three parallel encoders to capture spatial and morphological context at different resolutions. MERGE \cite{MERGE} constructs a hierarchical graph to model dependencies among morphologically similar but spatially distant clusters, enabling long-range node interactions.


Despite progress in multi-scale modeling, existing methods predict gene expression at the spot level and rely on datasets with fixed resolutions and uniform spot sizes, limiting their generalizability to heterogeneous spatial scales and geometries. To address this, we propose PixNet, a framework that predicts spatial gene expression at arbitrary resolutions and scales directly from histopathology images.

\paragraph{Dense Prediction in Histopathology and Medical Imaging. }

\begin{figure*}[t]
    \centering
    \begin{tikzpicture}
    \tikzset{font=\large}
    \node[anchor=south west,inner sep=0] (image) at (0,0) {\includegraphics[width=\linewidth]{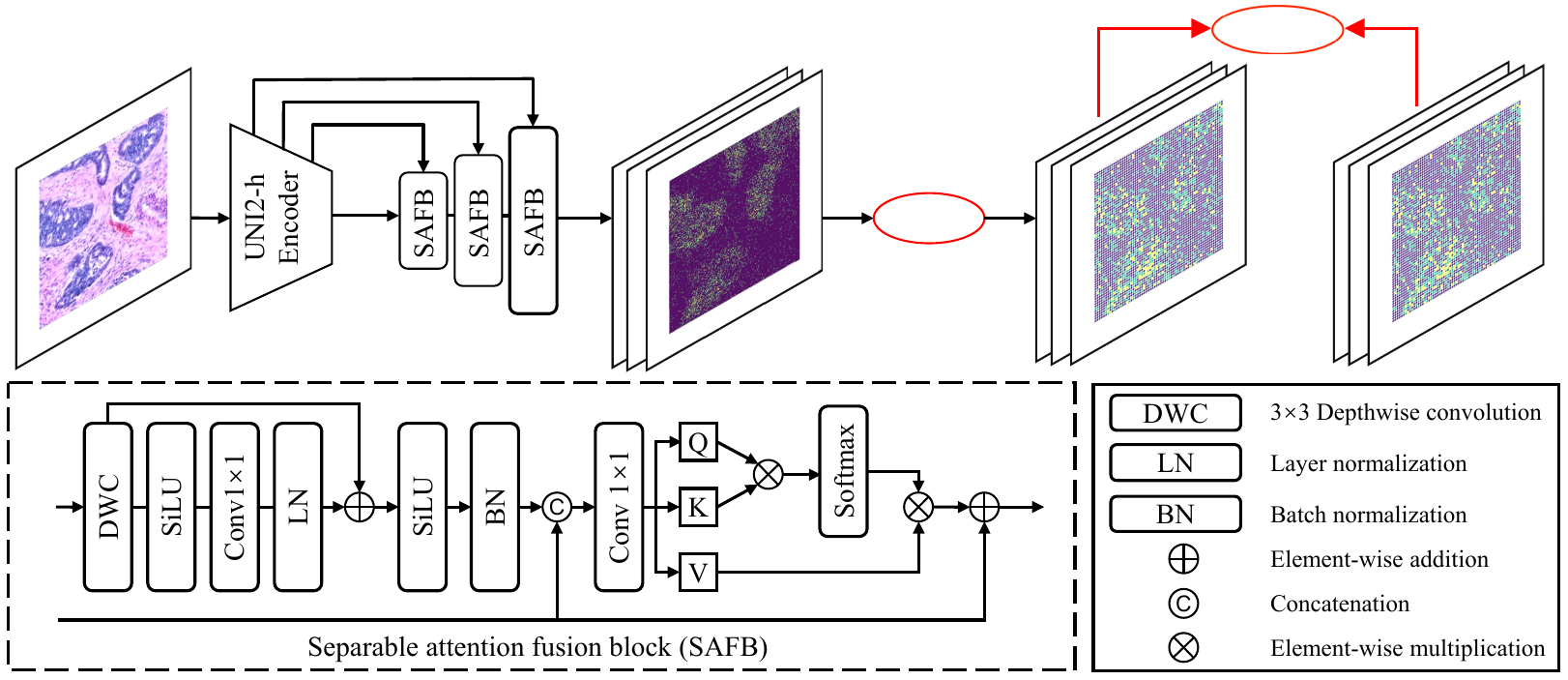}};
        \begin{scope}[x={(image.south east)},y={(image.north west)}, font=\small]
            \draw (0.025, 0.25) node {$\mathbf{F}_{l}$};
            \draw (0.025, 0.11) node {$\mathbf{U}_{l}$};
            \draw (0.025, 0.08) node[rotate=90] {$/$};
            \draw (0.025, 0.05) node {$\mathbf{\hat{U}}_{l}$};
            \draw (0.594, 0.67) node {Eq. 9};
            \draw (0.814, 0.96) node {$\mathcal{L}$};
            \draw (0.09, 0.51) node {$\mathbf{I}$};
            \draw (0.49, 0.51) node {$\mathbf{G}$};
            \draw (0.78, 0.51) node {$\{\hat{y}_{n}\}_{n=1}^{N}$};
            \draw (0.97, 0.51) node {$\{y_{n}\}_{n=1}^{N}$};
        \end{scope}
    \end{tikzpicture}
    \caption{Overview of our framework. 
    We extract a pyramidal feature map $\{\mathbf{F}_{l}\}_{l=1}^{L}$ from a slide image $\mathbf{I}$ and progressively decode them into a gene expression map $\mathbf{G}$ using multiple separable attention fusion blocks (SAFBs).
    The predicted gene expression values $\{\hat{y}_{n}\}_{n=1}^{N}$ are aggregated (Eq. \ref{eq_sum}) from $\mathbf{G}$ based on the positions and radiuses of spots of interest. The loss $\mathcal{L}$ is computed sparsely on spots with ground truth gene expression $\{y_{n}\}_{n=1}^{N}$ during training.
    } 
    \label{fig:backbone}
\end{figure*}


Recent advances in pixel-level dense prediction—particularly in medical image segmentation \cite{H-denseformer,U-Netmer,SelfReg‑UNet} and super-resolution \cite{Deform‑Mamba,istar}—have laid a technical foundation for constructing dense spatial gene expression maps from histopathology images. H-DenseFormer \cite{H-denseformer} combines CNNs and Transformers to achieve accurate segmentation with low complexity. U-Netmer \cite{U-Netmer} integrates Transformer-based self-attention into U-Net to address feature degradation and scale sensitivity. SelfReg-UNet \cite{SelfReg‑UNet} enhances U-Net's robustness via adaptive feature recalibration. Deform-Mamba \cite{Deform‑Mamba} uses deformable encoding and multi-view context learning to reconstruct high-quality MRI from low-resolution input, preserving fine structural details.

Dense prediction networks remain underexplored for spatial gene expression. Existing methods typically extract fixed-size patches (spots) to predict gene expression, causing loss of spatial resolution and limiting multi-scale prediction. iSTar \cite{istar} pioneers pixel-wise mapping from WSIs to gene expression, enabling high-resolution predictions. However, it focuses on super-resolution—inferring fine details from coarse data—which is ill-posed \cite{sr_survey} and often yields predictions inconsistent with true tissue morphology.


In this work, we directly predict gene expression from histology images without ill-posed upsampling, avoiding challenges of super-resolution and fixed spot sizes. This allows spatial gene expression prediction at arbitrary scales and resolutions.

\section{Method}

\paragraph{Overview.} Given a slide image $\mathbf{I} \in \mathbb{R}^{H \times W \times 3}$ and $N$ spots $\{x_{n},y_{n}\}_{n=1}^{N}$, each spot is centered at location $(x_n, y_n)$ on the histopathology slide image with a circular radius $r_{n}$ and is paired with expression $\mathbf{y}_{n} \in \mathbb{R}^{M}$ for $M$ genes. $H$ and $W$ are height and width of the slide image. We train a network to predict the gene expression with three steps:
1) the slide image is encoded into $L$ levels of pyramidal feature map $\{\mathbf{F}_{l}\}_{l=1}^{L}$; 
2) the pyramidal feature map $\{\mathbf{F}_{l}\}_{l=1}^{L}$ is decoded propressively into a dense gene expression map $\mathbf{G} \in \mathbb{R}^{H \times W \times M}$;
3) the gene expression $\hat{\mathbf{y}}_{n}$ is predicted for the spot centered at $(x_n, y_n)$ by summing over the values within a circular region of radius $r_{n}$ in the dense gene expression map $\mathbf{G}$.
During training, $\{\hat{\mathbf{y}}_{n}\}_{n=1}^{N}$ is optimized with the ground truth gene expression $\{\mathbf{y}\}_{n=1}^{N}$, forming our sparse loss module. During testing, the spot location $(x_n', y_n')$ and radius $r_{n}'$ can be varied dynamically, allowing for flexible gene expression prediction across different spatial resolutions and spot sizes. Figure \ref{fig:backbone} provides an overview of the proposed method.

\subsection{Pyramidal Feature Extraction}


To capture the features with varying scales and sizes in the slide image $\mathbf{I}$, we extract a pyramidal feature map $\{\mathbf{F}_{l}\}_{l=1}^{L}$ using UNI2-h \cite{UNI} encoder pretrained on a large WSI dataset. The encoder first projects $\mathbf{I}$ into token embeddings $\mathbf{Z}_0$, The embeddings are then processed by $L$ sequential vision transformer (ViT) groups to progressively enrich their semantic abstraction:
\begin{align}
    \mathbf{Z}_L = \text{ViT}_{L} \circ \text{ViT}_{L-1} \circ \dots \circ \text{ViT}_{1}(\mathbf{Z}_0),
\end{align}
where $\text{ViT}_{l}(\cdot)$ denotes the $l$-th vision transformer group.
Then we generate spatially pyramidal feature maps from intermediate transformer groups (e.g., from 2th, 4th, 6th transformer groups): $\mathbf{F}_{l} = \text{Downsample}\big(\mathcal{R}(\mathbf{Z}_l)\big)$, where $\mathcal{R}(\cdot)$ converts the sequence of tokens back to a 2D feature map.

\subsection{Gene Expression Map Decoding}

We faciliate the interactions among pyramidal feature map $\{\mathbf{F}_{l}\}_{l=1}^{L}$ to progressively decode a gene expression map $\mathbf{G}$. Our decoder network has $L-1$ layers, and follows an u-net style. Setting $\mathbf{F}_{L}$ as the initial decoder feature $\mathbf{D}_{L}$, we iteratively decode the pyramidal feature map $\{\mathbf{F}_{l}\}_{l=1}^{L}$. 
In deep stages we adopt a \textit{depth-to-space upsampling block} (DSUB), which rearranges feature maps into higher spatial resolution without altering their content. This design aims to minimize information loss, because in the task of spatial gene expression prediction, better preservation of spatial information learned by the encoder often leads to improved prediction performance \cite{triplex}.

Specifically, given an input feature map $\mathbf{F}_{l}$, the number of filters required for the upsampling operation is computed as: $K = C_{{\mathbf{F}_{l}}} \times 2^d$, where $C_{{\mathbf{F}_{l}}}$ denotes the number of channels in current feature map $\mathbf{F}_{l}$, and $d = 2$ represents the downsampling factor. The depth-to-space (D2S) \cite{D2S} operation then reshapes the channel dimension into higher spatial resolution. The overall process can be formulated as:
\begin{equation}
    \mathbf{U}_{l-1}=\text{CB}\Big(\text{ReLU}\big(\text{Conv}\big(\text{D2S}(\text{ReLU}(\text{Conv}(\mathbf{F}_{l}))), K\big)\big)\Big) \ ,
\end{equation}
where $\mathbf{U}_{l-1}$ denotes the upsampled feature map, $\text{CB}$ represents an composite convolutional block, and $\text{Conv}$ denotes a $3 \times 3$ convolutional layer.


\begin{table*}[!t]
    \centering
    \small 
    \caption{Quantitative gene expression prediction comparisons with state-of-the-art methods on two Visium HD datasets. The best-performing method is highlighted in bold.}
    \setlength{\tabcolsep}{17pt}
    \begin{tabular}{lccccc}
        \toprule
        Method  & MSE~$(\downarrow)$ & MAE~$(\downarrow)$  &PCC@F~$(\uparrow)$ &PCC@S~$(\uparrow)$ &PCC@M~$(\uparrow)$  \\
        \midrule
        \rowcolor{orange!10!white}
        \multicolumn{6}{l}{Experiments on the breast cancer Visium HD dataset.} \\
            STNet \cite{stnet} & 0.269$\scriptstyle \pm0.03$  & 0.482$\scriptstyle \pm0.05$  & -0.009$\scriptstyle \pm0.05$  & -0.001$\scriptstyle \pm0.06$  & -0.001$\scriptstyle \pm0.06$ \\
            HistoGene \cite{histogene}  & 0.265$\scriptstyle \pm0.02$  & 0.438$\scriptstyle \pm0.05$  & 0.089$\scriptstyle \pm0.07$  & 0.159$\scriptstyle \pm0.10$  & 0.157$\scriptstyle \pm0.10$ \\
            EGN \cite{egn}  & 0.264$\scriptstyle \pm0.03$  & 0.413$\scriptstyle \pm0.03$  & 0.102$\scriptstyle \pm0.03$ & 0.166$\scriptstyle \pm0.10$  & 0.161$\scriptstyle \pm0.08$ \\
            EGGN \cite{eggn2023}  & 0.241$\scriptstyle \pm0.02$  & 0.423$\scriptstyle \pm0.05$  & 0.143$\scriptstyle \pm0.02$  & 0.209$\scriptstyle \pm0.05$  & 0.200$\scriptstyle \pm0.06$ \\
            BLEEP \cite{bleep}  & 0.247$\scriptstyle \pm0.05$  & 0.435$\scriptstyle \pm0.05$  & 0.157$\scriptstyle \pm0.03$  & 0.216$\scriptstyle \pm0.08$  & 0.205$\scriptstyle \pm0.08$ \\
            HGGEP \cite{HGGEP}  & 0.267$\scriptstyle \pm0.10$ & 0.446$\scriptstyle \pm0.12$ & 0.118$\scriptstyle \pm0.03$ & 0.191$\scriptstyle \pm0.03$ & 0.187$\scriptstyle \pm0.04$  \\
            Junayed, et al. \cite{junayed}  & 0.306$\scriptstyle \pm0.06$ & 0.511$\scriptstyle \pm0.08$ & 0.047$\scriptstyle \pm0.10$ & 0.133$\scriptstyle \pm0.12$ & 0.134$\scriptstyle \pm0.12$  \\
            TRIPLEX \cite{triplex}  & 0.259$\scriptstyle \pm0.04$ & 0.432$\scriptstyle \pm0.08$ & 0.122$\scriptstyle \pm0.03$ & 0.200$\scriptstyle \pm0.04$ & 0.199$\scriptstyle \pm0.07$  \\
            iStar \cite{MERGE}  & 0.285$\scriptstyle \pm0.03$ & 0.479$\scriptstyle \pm0.06$ & 0.119$\scriptstyle \pm0.05$ & 0.213$\scriptstyle \pm0.02$ & 0.216$\scriptstyle \pm0.03$  \\
            SGN \cite{sgn}  & 0.230$\scriptstyle \pm0.03$ & 0.358$\scriptstyle \pm0.06$ & 0.173$\scriptstyle \pm0.07$ & 0.227$\scriptstyle \pm0.04$ & 0.226$\scriptstyle \pm0.02$  \\
            BG-TRIPLEX \cite{BG-TRIPLEX}  & 0.255$\scriptstyle \pm0.03$ & 0.448$\scriptstyle \pm0.07$ & 0.109$\scriptstyle \pm0.03$ & 0.187$\scriptstyle \pm0.04$ & 0.190$\scriptstyle \pm0.06$  \\
            ScstGCN \cite{ScstGCN}  & 0.229$\scriptstyle \pm0.03$ & 0.352$\scriptstyle \pm0.04$ & 0.159$\scriptstyle \pm0.08$ & 0.231$\scriptstyle \pm0.10$ & 0.225$\scriptstyle \pm0.14$  \\
            MERGE \cite{MERGE}  & 0.290$\scriptstyle \pm0.06$ & 0.458$\scriptstyle \pm0.03$ & 0.125$\scriptstyle \pm0.07$ & 0.227$\scriptstyle \pm0.04$ & 0.221$\scriptstyle \pm0.02$  \\
            Ours  & \textbf{0.153}$\scriptstyle \pm0.02$ & \textbf{0.274}$\scriptstyle \pm0.05$ & \textbf{0.196}$\scriptstyle \pm0.02$ & \textbf{0.313}$\scriptstyle \pm0.03$ & \textbf{0.325}$\scriptstyle \pm0.02$\\
         \midrule
         \rowcolor{orange!10!white}
         \multicolumn{6}{l}{Experiments on the brain cancer Visium HD dataset.} \\
            STNet \cite{stnet} & 0.282$\scriptstyle \pm0.03$  & 0.514$\scriptstyle \pm0.05$  & -0.002$\scriptstyle \pm0.05$  & 0.046$\scriptstyle \pm0.06$  & 0.044$\scriptstyle \pm0.06$ \\
            HistoGene \cite{histogene}  & 0.279$\scriptstyle \pm0.02$  & 0.457$\scriptstyle \pm0.05$  & 0.100$\scriptstyle \pm0.07$  & 0.171$\scriptstyle \pm0.10$  & 0.166$\scriptstyle \pm0.10$ \\
            EGN \cite{egn}  & 0.277$\scriptstyle \pm0.03$  & 0.463$\scriptstyle \pm0.03$  & 0.096$\scriptstyle \pm0.03$ & 0.177$\scriptstyle \pm0.10$  & 0.171$\scriptstyle \pm0.08$ \\
            EGGN \cite{eggn2023}  & 0.259$\scriptstyle \pm0.02$  & 0.459$\scriptstyle \pm0.05$  & 0.131$\scriptstyle \pm0.02$  & 0.195$\scriptstyle \pm0.05$  & 0.191$\scriptstyle \pm0.06$ \\
            BLEEP \cite{bleep}  & 0.243$\scriptstyle \pm0.05$  & 0.477$\scriptstyle \pm0.05$  & 0.121$\scriptstyle \pm0.03$  & 0.189$\scriptstyle \pm0.08$  & 0.186$\scriptstyle \pm0.08$ \\
            HGGEP \cite{HGGEP}  & 0.301$\scriptstyle \pm0.06$ & 0.476$\scriptstyle \pm0.04$ & 0.100$\scriptstyle \pm0.07$ & 0.168$\scriptstyle \pm0.10$ & 0.170$\scriptstyle \pm0.12$  \\
            Junayed, et al. \cite{junayed}  & 0.335$\scriptstyle \pm0.06$ & 0.549$\scriptstyle \pm0.08$ & 0.083$\scriptstyle \pm0.07$ & 0.119$\scriptstyle \pm0.07$ & 0.110$\scriptstyle \pm0.08$  \\
            TRIPLEX \cite{triplex}  & 0.257$\scriptstyle \pm0.04$ & 0.462$\scriptstyle \pm0.08$ & 0.132$\scriptstyle \pm0.03$ & 0.195$\scriptstyle \pm0.04$ & 0.188$\scriptstyle \pm0.07$  \\
            iStar \cite{MERGE}  & 0.304$\scriptstyle \pm0.03$ & 0.481$\scriptstyle \pm0.06$ & 0.118$\scriptstyle \pm0.05$ & 0.190$\scriptstyle \pm0.02$ & 0.191$\scriptstyle \pm0.03$  \\
            SGN \cite{sgn}  & 0.228$\scriptstyle \pm0.03$ & 0.447$\scriptstyle \pm0.06$ & 0.133$\scriptstyle \pm0.07$ & 0.199$\scriptstyle \pm0.04$ & 0.195$\scriptstyle \pm0.02$  \\
            BG-TRIPLEX \cite{BG-TRIPLEX}  & 0.255$\scriptstyle \pm0.02$ & 0.446$\scriptstyle \pm0.04$ & 0.136$\scriptstyle \pm0.05$ & 0.191$\scriptstyle \pm0.07$ & 0.190$\scriptstyle \pm0.07$  \\
            ScstGCN \cite{ScstGCN}  & 0.242$\scriptstyle \pm0.03$ & 0.468$\scriptstyle \pm0.05$ & 0.117$\scriptstyle \pm0.05$ & 0.183$\scriptstyle \pm0.04$ & 0.185$\scriptstyle \pm0.06$  \\
            MERGE \cite{MERGE}  & 0.274$\scriptstyle \pm0.03$ & 0.501$\scriptstyle \pm0.06$ & 0.126$\scriptstyle \pm0.07$ & 0.187$\scriptstyle \pm0.04$ & 0.190$\scriptstyle \pm0.02$  \\
            Ours  & \textbf{0.144}$\scriptstyle \pm0.02$ & \textbf{0.298}$\scriptstyle \pm0.05$ & \textbf{0.176}$\scriptstyle \pm0.02$ & \textbf{0.305}$\scriptstyle \pm0.03$ & \textbf{0.304}$\scriptstyle \pm0.02$\\   
        \bottomrule
    \end{tabular}
    \label{tab:benchmark0}
\end{table*}

In the shallow stages, we adopt bilinear interpolation instead of the depth-to-space operation for upsampling, as it better preserves spatial details. Since shallow features retain rich spatial information, they mainly benefit from resolution recovery rather than complex semantic transformation. This operation ensures smoother reconstruction and maintains spatial coherence, which is essential for accurate spatial gene expression prediction. Prior to interpolation, a composite convolutional block is applied to suppress noise from non-informative features. A second composite block is then applied after interpolation to further refine the feature map. This process can be formally represented as:
\begin{equation}
    \mathbf{\hat{U}}_{l-1}=\text{CB}\Big(\text{BlIntp}\big(\text{CB}(\mathbf{F}_{l}) \big)\Big) \ ,
\end{equation}
where $\mathbf{\hat{U}}_{l-1}$ denotes the upsampled feature map in the shallow stages, and $\text{BlIntp}$ donotes bilinear interpolation upsampling operation.

Then we adopt a \textit{separable attention fusion block} (SAFB) to refine and aggregate the pyramidal feature map $\mathbf{F}_{l-1}$ and the upsampled feature map $\mathbf{U}_{l-1}/\mathbf{\hat{U}}_{l-1}$. Specifically, a specialized residual block is first applied to refine the pyramidal feature map $\mathbf{F}_{l-1}$. This process can be formulated as:
\begin{align}
    &\mathbf{\hat{F}}_\text{mid} = \text{LN}\Big(\text{Conv}_{1 \times 1}\big(\text{SiLU}(\text{DWC}(\mathbf{F}_{l-1}))\big)\Big) \ ,  \\
    &\mathbf{\hat{F}}_{l-1} = \text{BN}\Big(\text{SiLU}\big(\text{DWC}(\mathbf{F}_{l-1}) + \mathbf{\hat{F}}_\text{mid}\big)\Big) \ , 
\end{align}
where $\mathbf{\hat{F}}_{l-1}$ denotes the refined pyramidal feature map, $\mathbf{\hat{F}}_\text{mid}$ denotes the intermediate feature map, $\text{LN}$, $\text{BN}$, $\text{SiLU}$, $\text{DWC}$ denote layer normalization, batch normalization, Sigmoid linear unit, and depthwise convolution, respectively.

Next, we concatenate the refined pyramidal output $\mathbf{\hat{F}}_{l-1}$ with the upsampled feature map $\mathbf{U}_{l-1}/\mathbf{\hat{U}}_{l-1}$ to fuse high-resolution details from the encoder and contextual information from the decoder. A linear projection is applied to generate queries ($\mathbf{Q}$), keys ($\mathbf{K}$), and values ($\mathbf{V}$), followed by the computation of attention weights to enhance feature representations. The entire process can be formulated as: 
\begin{align}
    &\text{Attention} = \text{softmax}\left(\frac{\mathbf{Q}\mathbf{K}^\text{T}}{\beta}\right) \mathbf{V}  \ , \\
    & \mathbf{F}_\text{u} = \begin{cases}
        \text{Concat}(\mathbf{\hat{F}}_{l-1}, \mathbf{U}_{l-1}) \ , & \text{for } L\geq l > 3 \\
        \text{Cancat}(\mathbf{\hat{F}}_{l-1}, \mathbf{\hat{U}}_{l-1}) \ ,  & \text{for } 3\geq l > 1 \\
    \end{cases}     \\
    &\mathbf{D}_{l-1} = \mathbf{F}_\text{u} + \text{Attention}\big(\text{Conv}_{1 \times 1}(\mathbf{F}_\text{u})\big) \ ,
\end{align}
where $\beta$ is a normalization factor that is typically set to the square root of the latent dimension, and $\text{BI}$ donotes bilinear interpolation upsampling.

\begin{table*}[!t]
    \centering
    \small 
    \caption{Quantitative gene expression prediction comparisons with state-of-the-art methods on STNet and Her2ST datasets. The best-performing method is highlighted in bold.}
    \setlength{\tabcolsep}{17pt}
    \begin{tabular}{lccccc}
        \toprule
        Method  & MSE~$(\downarrow)$ & MAE~$(\downarrow)$  &PCC@F~$(\uparrow)$ &PCC@S~$(\uparrow)$ &PCC@M~$(\uparrow)$  \\
        \midrule
        \rowcolor{orange!10!white}
        \multicolumn{6}{l}{Experiments on the STNet~\cite{stnet} dataset.} \\
            STNet \cite{stnet} & 0.209$\scriptstyle \pm0.02$ & 0.502$\scriptstyle \pm0.05$ & 0.005$\scriptstyle \pm0.06$ & 0.092$\scriptstyle \pm0.07$ & 0.093$\scriptstyle \pm0.06$ \\
            HistoGene \cite{histogene} & 0.194$\scriptstyle \pm0.09$ & 0.418$\scriptstyle \pm0.12$ & 0.097$\scriptstyle \pm0.10$ & 0.126$\scriptstyle \pm0.11$ & 0.119$\scriptstyle \pm0.12$ \\
           EGN \cite{egn} & 0.192$\scriptstyle \pm0.02$ & 0.449$\scriptstyle \pm0.04$ & 0.106$\scriptstyle \pm0.05$ & 0.221$\scriptstyle \pm0.07$ & 0.203$\scriptstyle \pm0.09$ \\
           EGGN \cite{eggn2023} & 0.189$\scriptstyle \pm0.03$ & 0.424$\scriptstyle \pm0.06$ & 0.184$\scriptstyle \pm0.05$ & 0.305$\scriptstyle \pm0.05$ & 0.292$\scriptstyle \pm0.06$ \\
           BLEEP \cite{bleep} & 0.235$\scriptstyle \pm0.02$ & 0.451$\scriptstyle \pm0.05$ & 0.155$\scriptstyle \pm0.05$ & 0.208$\scriptstyle \pm0.05$ & 0.193$\scriptstyle \pm0.10$ \\
           HGGEP \cite{HGGEP}  & 0.212$\scriptstyle \pm0.04$ & 0.406$\scriptstyle \pm0.05$ & 0.137$\scriptstyle \pm0.07$ & 0.265$\scriptstyle \pm0.07$ & 0.253$\scriptstyle \pm0.08$  \\
           Junayed, et al. \cite{junayed}  & 0.239$\scriptstyle \pm0.02$ & 0.472$\scriptstyle \pm0.04$ & 0.138$\scriptstyle \pm0.04$ & 0.195$\scriptstyle \pm0.05$ & 0.190$\scriptstyle \pm0.05$  \\
           TRIPLEX \cite{triplex} & 0.202$\scriptstyle \pm0.02$ & 0.413$\scriptstyle \pm0.03$ & 0.159$\scriptstyle \pm0.04$ & 0.364$\scriptstyle \pm0.05$ & 0.352$\scriptstyle \pm0.10$ \\
           iStar \cite{MERGE}  & 0.241$\scriptstyle \pm0.03$ & 0.507$\scriptstyle \pm0.06$ & 0.155$\scriptstyle \pm0.05$ & 0.312$\scriptstyle \pm0.02$ & 0.320$\scriptstyle \pm0.03$  \\
           SGN \cite{sgn}  & 0.186$\scriptstyle \pm0.02$ & 0.388$\scriptstyle \pm0.05$ & 0.179$\scriptstyle \pm0.05$ & 0.289$\scriptstyle \pm0.06$ & 0.269$\scriptstyle \pm0.07$ \\
           BG-TRIPLEX \cite{BG-TRIPLEX}  & 0.201$\scriptstyle \pm0.03$ & 0.415$\scriptstyle \pm0.06$ & 0.162$\scriptstyle \pm0.07$ & 0.366$\scriptstyle \pm0.04$ & 0.357$\scriptstyle \pm0.02$  \\
           ScstGCN \cite{ScstGCN}  & 0.223$\scriptstyle \pm0.03$ & 0.405$\scriptstyle \pm0.03$ & 0.158$\scriptstyle \pm0.06$ & 0.337$\scriptstyle \pm0.16$ & 0.325$\scriptstyle \pm0.14$  \\
           MERGE \cite{MERGE}  & 0.195$\scriptstyle \pm0.03$ & 0.396$\scriptstyle \pm0.06$ & 0.153$\scriptstyle \pm0.03$ & 0.351$\scriptstyle \pm0.05$ & 0.344$\scriptstyle \pm0.05$  \\
           Ours & \textbf{0.146}$\scriptstyle \pm0.04$ & \textbf{0.301}$\scriptstyle \pm0.04$ & \textbf{0.207}$\scriptstyle \pm0.03$ & \textbf{0.416}$\scriptstyle \pm0.02$ & \textbf{0.409}$\scriptstyle \pm0.05$ \\
        \midrule
         \rowcolor{orange!10!white}
         \multicolumn{6}{l}{Experiments on the Her2ST~\cite{her2st} dataset.} \\
            STNet \cite{stnet} & 0.260$\scriptstyle \pm0.04$  & 0.491$\scriptstyle \pm0.07$  & 0.093$\scriptstyle \pm0.05$  & 0.126$\scriptstyle \pm0.06$  & 0.145$\scriptstyle \pm0.16$ \\
            HistoGene \cite{histogene}  & 0.314$\scriptstyle \pm0.09$  & 0.531$\scriptstyle \pm0.05$  & 0.128$\scriptstyle \pm0.07$  & 0.297$\scriptstyle \pm0.15$  & 0.302$\scriptstyle \pm0.17$ \\
            EGN \cite{egn}  & 0.241$\scriptstyle \pm0.06$  & 0.464$\scriptstyle \pm0.03$  & 0.129$\scriptstyle \pm0.03$ & 0.330$\scriptstyle \pm0.17$  & 0.328$\scriptstyle \pm0.17$ \\
            EGGN \cite{eggn2023}  & 0.259$\scriptstyle \pm0.02$  & 0.459$\scriptstyle \pm0.05$  & 0.131$\scriptstyle \pm0.02$  & 0.332$\scriptstyle \pm0.05$  & 0.326$\scriptstyle \pm0.06$ \\
            BLEEP \cite{bleep}  & 0.277$\scriptstyle \pm0.05$  & 0.498$\scriptstyle \pm0.05$  & 0.106$\scriptstyle \pm0.03$  & 0.285$\scriptstyle \pm0.08$  & 0.277$\scriptstyle \pm0.08$ \\
            HGGEP \cite{HGGEP}  & 0.312$\scriptstyle \pm0.02$ & 0.547$\scriptstyle \pm0.05$ & 0.100$\scriptstyle \pm0.10$ & 0.244$\scriptstyle \pm0.08$ & 0.239$\scriptstyle \pm0.08$  \\
            Junayed, et al. \cite{junayed}  & 0.315$\scriptstyle \pm0.05$ & 0.532$\scriptstyle \pm0.07$ & 0.099$\scriptstyle \pm0.15$ & 0.284$\scriptstyle \pm0.09$ & 0.280$\scriptstyle \pm0.12$  \\
            TRIPLEX \cite{triplex}  & 0.226$\scriptstyle \pm0.04$ & 0.462$\scriptstyle \pm0.08$ & 0.151$\scriptstyle \pm0.03$ & 0.415$\scriptstyle \pm0.04$ & 0.404$\scriptstyle \pm0.07$  \\
            iStar \cite{MERGE}  & 0.296$\scriptstyle \pm0.03$ & 0.541$\scriptstyle \pm0.06$ & 0.138$\scriptstyle \pm0.05$ & 0.336$\scriptstyle \pm0.02$ & 0.347$\scriptstyle \pm0.03$  \\
            SGN \cite{sgn}  & 0.247$\scriptstyle \pm0.03$ & 0.465$\scriptstyle \pm0.06$ & 0.117$\scriptstyle \pm0.07$ & 0.360$\scriptstyle \pm0.04$ & 0.355$\scriptstyle \pm0.02$  \\
            BG-TRIPLEX \cite{BG-TRIPLEX}  & 0.238$\scriptstyle \pm0.04$ & 0.478$\scriptstyle \pm0.05$ & 0.135$\scriptstyle \pm0.16$ & 0.358$\scriptstyle \pm0.09$ & 0.344$\scriptstyle \pm0.11$  \\
            ScstGCN \cite{ScstGCN}  & 0.301$\scriptstyle \pm0.02$ & 0.506$\scriptstyle \pm0.04$ & 0.094$\scriptstyle \pm0.05$ & 0.295$\scriptstyle \pm0.04$ & 0.290$\scriptstyle \pm0.02$  \\
            MERGE \cite{MERGE}  & 0.244$\scriptstyle \pm0.03$ & 0.481$\scriptstyle \pm0.03$ & 0.130$\scriptstyle \pm0.07$ & 0.349$\scriptstyle \pm0.12$ & 0.342$\scriptstyle \pm0.10$  \\
            Ours  & \textbf{0.199}$\scriptstyle \pm0.02$ & \textbf{0.368}$\scriptstyle \pm0.04$ & \textbf{0.185}$\scriptstyle \pm0.02$ & \textbf{0.452}$\scriptstyle \pm0.03$ & \textbf{0.453}$\scriptstyle \pm0.03$\\
         \bottomrule 
    \end{tabular}
    \label{tab:benchmark1}
\end{table*}


For the last feature map $\mathbf{D}_{1}$, we apply a $1{\times}1$ convolution produce the gene expression map $\mathbf{G}.$

 
\subsection{Gene Expression Prediction}
For a spot with location $(x_{n},y_{n})$ and radius $r_{n}$ on the slide image $\mathbf{I}$, the predicted gene expression $\hat{\mathbf{y}}_{n}$ is 
\begin{align}
    \hat{\mathbf{y}}_{n} = \sum_{\substack{\Delta x, \Delta y}} \mathbf{G}(\Delta x, \Delta y) \ , \label{eq_sum}
\end{align}
where the sum enumeration all postions $(\Delta x, \Delta y)$ on the domain of $(\Delta x - x_{n} )^2 + ( \Delta y - y_{n} )^2 \leq r_n^2$, and $\mathbf{G}(\Delta x, \Delta y)$ denotes the predicted gene expression at position $(\Delta x, \Delta y)$. This aggregation effectively pools the gene expression values within the spot, yielding a robust prediction for each spot.

\subsection{Loss Function}
We sparsely supervise the predicted gene expression map $\mathbf{G}$ with the mean square error $\mathcal{L}_{\text{mse}}$ and batch-wise Pearson correlation coefficient (PCC) loss $\mathcal{L}_{\text{pcc}}$. The loss function penalizes penalizes deviations of aggregated gene expression $\{\hat{y}_{n}\}_{n=1}^{N}$ from the ground truth gene expression $\{y_{n}\}_{n=1}^{N}$, and encourages a correlation between them. The overall training loss $\mathcal{L}$ is 
\begin{equation}
    \mathcal{L} = \mathcal{L}_{\text{mse}} + \lambda \cdot \mathcal{L}_{\text{pcc}}  \ ,
\end{equation}
where $\lambda$ is a hyperparameter that controls the relative importance of the PCC loss compared to the mean square error.

\section{Experiment}

\paragraph{Datasets.}
We experiment with four common datasets:
1) STNet dataset~\cite{stnet} that has 68 slide images with 30K spots on 100 $\mu$m; 
2) Her2ST dataset~\cite{her2st} that has 36 slide images with 13K spots on 100 $\mu$m; 
3) Breast cancer Visium HD dataset, providing 2 slide images with 18.7M, 1.17M, and 294K spots on 2 $\mu$m, 8 $\mu$m, and 16 $\mu$m; 
4) Brain cancer Visium HD dataset with 14.2M, 889K, and 223K spots on 2 $\mu$m, 8 $\mu$m, and 16 $\mu$m from 2 slide images; 
We follow the preprocessing and cross-fold validation protocols outlined in~\cite{eggn2023,sgn}. The Visium HD dataset is downloaded from 10xProteomic, including the Visium HD Spatial Gene Expression Library, Human Breast Cancer; and Visium HD Spatial Gene Expression Library, Mouse Brain.

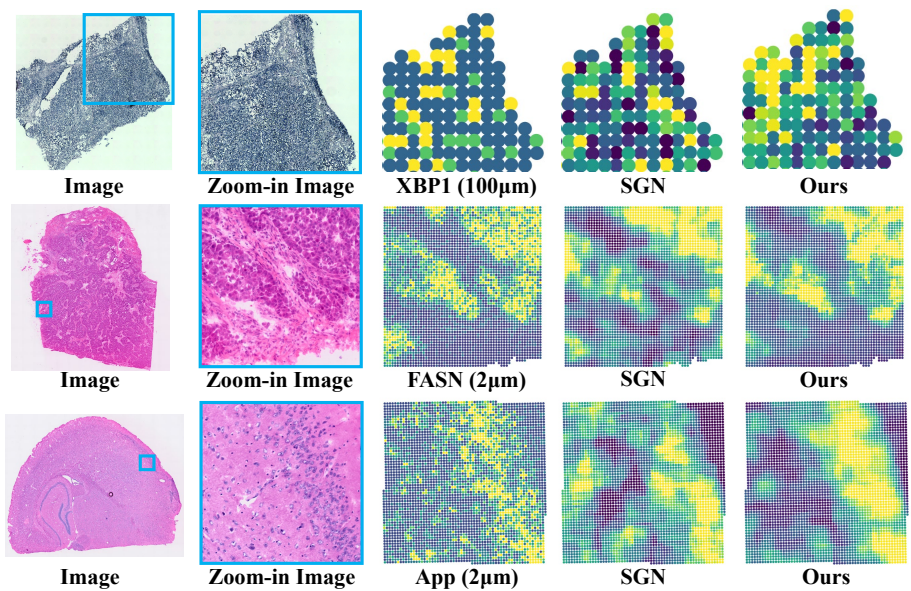
\begin{figure*}[t]
    \centering
    \begin{tikzpicture}
     \node[anchor=south west,inner sep=0] (image) at (0,0) {\includegraphics[width=\linewidth]{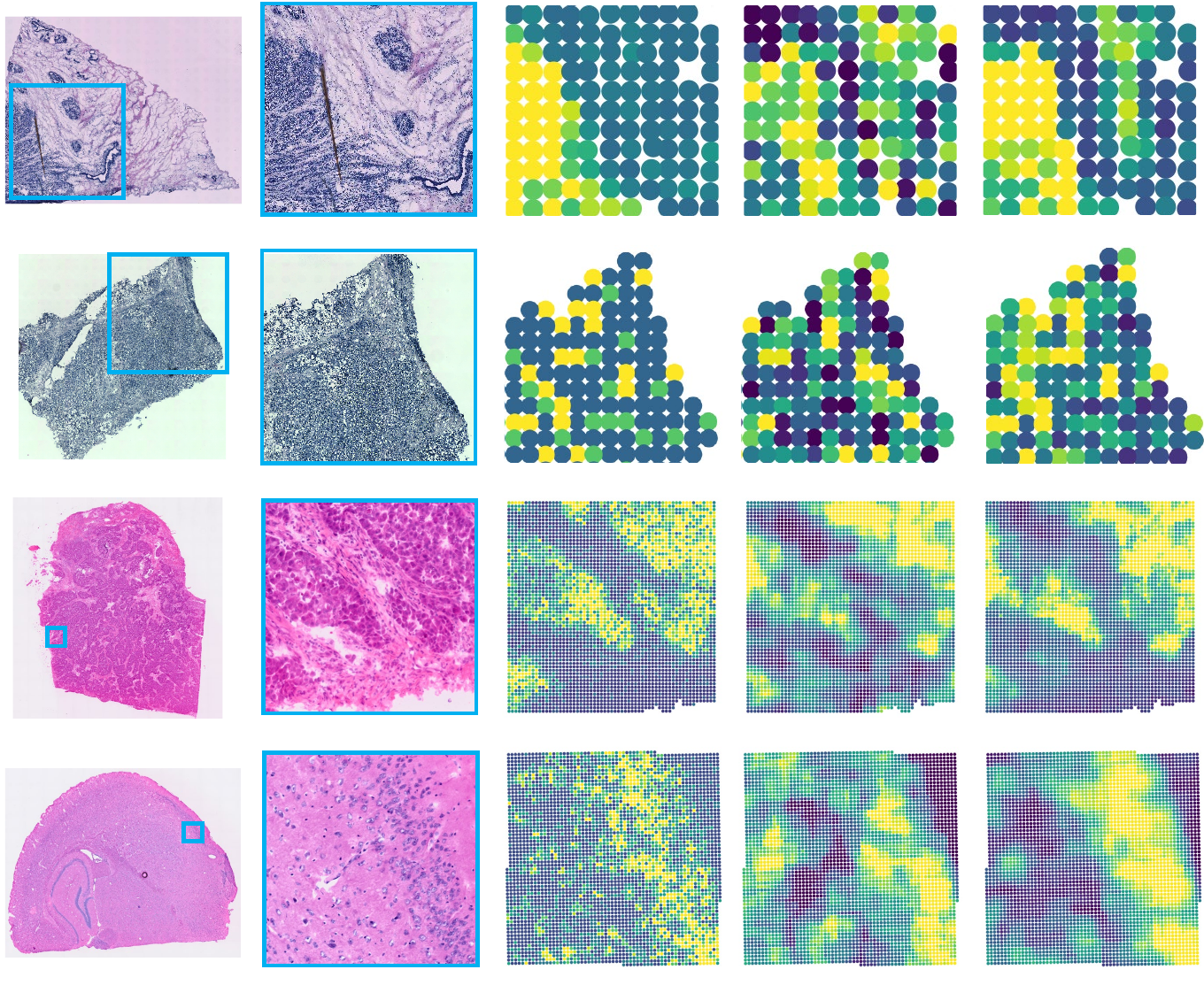}};
        \begin{scope}[x={(image.south east)},y={(image.north west)}]

            \draw (0.31, 0.025) node {Zoom-in Image};
            \draw (0.10, 0.025) node {Image};
            \draw (0.51 , 0.025) node {App (2 $\mu$m)};
            \draw (0.70 , 0.025) node {SGN};
            \draw (0.91 , 0.025) node {Ours};

            \draw (0.31, 0.275) node {Zoom-in Image};
            \draw (0.10, 0.275) node {Image};
            \draw (0.51 , 0.275) node {FASN (2 $\mu$m)};
            \draw (0.70 , 0.275) node {SGN};
            \draw (0.91 , 0.275) node {Ours};
            
            \draw (0.31, 0.525) node {Zoom-in Image};
            \draw (0.10, 0.525) node {Image};
            \draw (0.51 , 0.525) node {XBP1 (100 $\mu$m)};
            \draw (0.70 , 0.525) node {SGN};
            \draw (0.91 , 0.525) node {Ours};
            
            \draw (0.31, 0.77) node {Zoom-in Image};
            \draw (0.10, 0.77) node {Image};
            \draw (0.51 , 0.77) node {GNAS (100 $\mu$m)};
            \draw (0.70 , 0.77) node {SGN};
            \draw (0.91 , 0.77) node {Ours};
        \end{scope}
    \end{tikzpicture}
    \caption{Examples of predicted expression of gene types that are related to cancers \cite{stnet,App_0,App_1,XBP1_0,XBP1_1,FASN_0,FASN_1}. From left to right, we show the slide image, ground truth gene expression, and predictions from various methods are shown for regions cropped from the colored boxes in the slide image.
    } 
    \label{fig:visual}
\end{figure*}

\paragraph{Evaluation Metrics.}
Our method is evaluated using the following metrics: mean squared error (MSE), mean absolute error (MAE), first quartile of Pearson correlation coefficient (PCC@F), median of Pearson correlation coefficient (PCC@S), and mean of Pearson correlation coefficient (PCC@M).

\paragraph{Implementation Details.} Our method is fully implemented using Pytorch \cite{pyg,pytorch}, and all experiments are conducted on a single NVIDIA RTX A6000 GPU. Our method is trained entirely from scratch, with the AdamW \cite{adamw} optimizer for 200 epochs, using a learning rate of $5 \times 10^{-4}$ and a weight decay of $1 \times 10^{-4}$. The decoder is configured with standard filter sizes [64, 128, 256, 512, 512, 512] following common encoder–decoder design practice \cite{standard_encdec}. We set $\lambda$ to 0.5. A fixed random seed (42) is used across all implementations and reruns. Following the approach in \cite{stnet}, we select the 250 genes with the highest mean expression across the dataset as prediction targets. The gene expression values are normalized by dividing by the sum of expressions in each spot, followed by a log transformation \cite{triplex}. For reproducibility and fair comparison, each experiment is repeated five times, and the average score is reported. More implementation details can be found in the supplementary material.

\paragraph{Baseline Methods.}
We compare with 13 state-of-the-art (SOTA) spatial gene expression prediction methods \cite{stnet,histogene,egn,eggn2023,bleep,HGGEP,junayed,triplex,istar,sgn,BG-TRIPLEX,ScstGCN,MERGE}. To maintain evaluation consistency and reproducibility, most of the methods in Tab. \ref{tab:benchmark0} are retrained within the same local setup. A portion of the results in Tab. \ref{tab:benchmark1} are adapted from \cite{triplex}.

\subsection{Experimental Result}

We benchmark our method against state-of-the-art approaches on four ST datasets. Results on the Breast Cancer Visium HD and Brain Cancer Visium HD datasets are shown in Tab. \ref{tab:benchmark0}, while comparisons on the STNet \cite{stnet} and Her2ST \cite{her2st} datasets are shown in Tab. \ref{tab:benchmark1}.


Our method achieves state-of-the-art performance across all four ST datasets, demonstrating competitive results and, in particular, leading PCC-based performance. This is particularly important for gene expression prediction, as it reflects the model’s ability to capture relative variations in gene expression more effectively. For instance, on the breast cancer Visium HD dataset, our method achieves a PCC@M of 0.325—an 43.8\% improvement over the previous state-of-the-art method, SGN (0.226 PCC@M). Similarly, on the Her2ST dataset, our method achieves a PCC@M of 0.453—representing a 12.1\% improvement over the previous best method, TRIPLEX (0.404 PCC@M).


Fig. \ref{fig:visual} shows qualitative comparisons with competitive methods. Our method produces visual patterns that more closely resemble the ground truth gene expression at both large (100 $\mu$m) and single-cell (2 $\mu$m) spot sizes, demonstrating robustness across scales.

\begin{table}[h]
    \centering
    \small 
    \caption{Ablation study of our proposed SAFB module .}
    \setlength{\tabcolsep}{9pt}
    \begin{tabular}{lcccc}
    \toprule
      Decoder & MSE$(\downarrow)$ & MAE$(\downarrow)$ & PCC@M$(\uparrow)$ \\
     \midrule
      Conv &  0.368$\scriptstyle \pm0.04$ & 0.470$\scriptstyle \pm0.07$ & 0.169$\scriptstyle \pm0.13$ \\
      ResNet18 & 0.226$\scriptstyle \pm0.03$ & 0.374$\scriptstyle \pm0.05$ & 0.213$\scriptstyle \pm0.09$ \\
      ViT & 0.297$\scriptstyle \pm0.07$ & 0.435$\scriptstyle \pm0.12$ & 0.188$\scriptstyle \pm0.04$ \\
      \textbf{SAFB} & \textbf{0.153}$\scriptstyle \pm0.02$ & \textbf{0.274}$\scriptstyle \pm0.05$ & \textbf{0.325}$\scriptstyle \pm0.02$ \\
     \bottomrule
    \end{tabular}
    \label{tab:ablation_component}
\end{table}

\begin{table}[!t]
    \centering
    \small 
    \caption{Ablation study of gene expression measured for different loss functions used in training.}
    \setlength{\tabcolsep}{8pt}
    \begin{tabular}{ccccc}
    \toprule
      $\mathcal{L}_{\text{mse}}$ & $\mathcal{L}_{\text{pcc}}$ & MSE$(\downarrow)$ & MAE$(\downarrow)$ & PCC@M$(\uparrow)$ \\
     \midrule
      \ding{51} & \ding{55} & 0.170$\scriptstyle \pm0.06$ & 0.286$\scriptstyle \pm0.10$ & 0.293$\scriptstyle \pm0.03$ \\
      \ding{55} & \ding{51} & 0.192$\scriptstyle \pm0.09$ & 0.305$\scriptstyle \pm0.15$ & 0.319$\scriptstyle \pm0.08$ \\
      \ding{51} & \ding{51} & \textbf{0.153}$\scriptstyle \pm0.02$ & \textbf{0.274}$\scriptstyle \pm0.05$ & \textbf{0.325}$\scriptstyle \pm0.02$ \\
     \bottomrule
    \end{tabular}
    \label{tab:ablation_loss}
\end{table}

\begin{table*}[!t]
    \centering
    \small 
    \caption{Performance generalization. All models are trained on the STNet~\cite{stnet} dataset (with spot size $100 \mu m$), and tested on the breast cancer Visium HD dataset with varying spot sizes ($2 \mu m$, $8 \mu m$, and $16 \mu m$) and slide images from different environments. The standard deviation is not displayed due to space limitations. }
    \setlength{\tabcolsep}{4pt}
    \begin{tabular}{l|ccc|ccc|ccc}
        \toprule
        & \multicolumn{3}{c|}{$2 \mu m$}
        & \multicolumn{3}{c|}{$8 \mu m$}
        & \multicolumn{3}{c}{$16 \mu m$}\\
        \multirow{-2}*{Method}  & {MSE$(\downarrow)$} & {MAE$(\downarrow)$} & {PCC@M$(\uparrow)$} & {MSE$(\downarrow)$} & {MAE$(\downarrow)$} & {PCC@M$(\uparrow)$} & {MSE$(\downarrow)$} & {MAE$(\downarrow)$} & {PCC@M$(\uparrow)$} \\
         \midrule
           STNet \cite{stnet} & 0.420 & 0.515 & 0.000 & 0.397 & 0.498 & 0.002 & 0.395 & 0.501 & 0.005 \\
           HistoGene \cite{histogene} & 0.388 & 0.501 & 0.088 & 0.389 & 0.500 & 0.093 & 0.373 & 0.485 & 0.097 \\
           EGN \cite{egn} & 0.371 & 0.492 & 0.091 & 0.377 & 0.479 & 0.087 & 0.365 & 0.481 & 0.100 \\
           EGGN \cite{eggn2023} & 0.368 & 0.483 & 0.087 & 0.372 & 0.499 & 0.088 & 0.361 & 0.487 & 0.102 \\
           BLEEP \cite{bleep} & 0.323 & 0.441 & 0.115 & 0.302 & 0.431 & 0.127 & 0.301 & 0.435 & 0.125 \\
           HGGEP \cite{HGGEP} & 0.405 & 0.587 & 0.068 & 0.411 & 0.543 & 0.083 & 0.412 & 0.537 & 0.094 \\
           Junayed, et al. \cite{junayed} & 0.339 & 0.469 & 0.106 & 0.343 & 0.496 & 0.116 & 0.337 & 0.446 & 0.120 \\
           TRIPLEX \cite{triplex} & 0.365 & 0.488 & 0.092 & 0.355 & 0.451 & 0.100 & 0.356 & 0.457 & 0.099 \\
           iStar \cite{istar} & 0.327 & 0.453 & 0.126 & 0.324 & 0.439 & 0.133 & 0.344 & 0.480 & 0.137 \\
           SGN \cite{sgn}  & 0.303 & 0.437 &  0.118 & 0.289 & 0.422 & 0.136 & 0.286 & 0.431 & 0.123 \\
           BG-TRIPLEX \cite{BG-TRIPLEX} & 0.370 & 0.492 & 0.103 & 0.366 & 0.501 & 0.094 & 0.348 & 0.482 & 0.111 \\
           ScstGCN \cite{ScstGCN} & 0.346 & 0.445 & 0.109 & 0.341 & 0.450 & 0.112 & 0.351 & 0.459 & 0.107 \\
           MERGE \cite{MERGE} & 0.352 & 0.454 & 0.112 & 0.345 & 0.458 & 0.127 & 0.334 & 0.474 & 0.139 \\
           \textbf{Ours} & \textbf{0.215} & \textbf{0.382} & \textbf{0.198} & \textbf{0.197} & \textbf{0.370} & \textbf{0.219} & \textbf{0.199} & \textbf{0.365} & \textbf{0.226} \\
         \bottomrule         
    \end{tabular}
    \label{tab:gen}
\end{table*}

\begin{table}[!t]
    \centering
    \small 
    \caption{Ablation study of different spot sizes used in training.}
    \setlength{\tabcolsep}{4pt}
    \begin{tabular}{cccccc}
    \toprule
         \multicolumn{3}{c}{Spot size} &  \multirow{2}{*}{MSE$(\downarrow)$} & \multirow{2}{*}{MAE$(\downarrow)$} & \multirow{2}{*}{PCC@M$(\uparrow)$} \\
    16\,$\mu m$ & 8\,$\mu m$ & 2\,$\mu m$  & & &\\
     \midrule
      \ding{51} & \ding{55} & \ding{55} & 0.208$\scriptstyle \pm0.03$ & 0.297$\scriptstyle \pm0.04$ & 0.244$\scriptstyle \pm0.06$\\
      \ding{55} & \ding{51} & \ding{55} & 0.209$\scriptstyle \pm0.03$ & 0.302$\scriptstyle \pm0.05$ & 0.288$\scriptstyle \pm0.06$\\
      \ding{55} & \ding{55} & \ding{51} & 0.186$\scriptstyle \pm0.02$ & 0.291$\scriptstyle \pm0.04$ & 0.299$\scriptstyle \pm0.04$\\
      \ding{51} & \ding{51} & \ding{55} & 0.177$\scriptstyle \pm0.04$ & 0.285$\scriptstyle \pm0.06$ & 0.303$\scriptstyle \pm0.07$\\
      \ding{51} & \ding{51} & \ding{51} & \textbf{0.153}$\scriptstyle \pm0.02$ & \textbf{0.274}$\scriptstyle \pm0.05$ & \textbf{0.325}$\scriptstyle \pm0.02$ \\
     \bottomrule
    \end{tabular}
    \label{tab:ablation_scale}
\end{table}

\subsection{Ablation Study}

\paragraph{Model Architecture.} 
We compare several decoders for mapping pyramidal features to dense expression maps—plain Conv, ResNet18 \cite{resnet}, ViT \cite{vit}, and our SAFB module (Tab. \ref{tab:ablation_component}). SAFB delivers the best results across all metrics, demonstrating that its lightweight, relation-aware aggregation better preserves local fidelity and global consistency of expression hotspots.

\paragraph{Loss Function.} 
We evaluate three training objectives—$\mathcal{L}_{\text{mse}}$ only, $\mathcal{L}_{\text{pcc}}$ only, and their combination—on the breast-cancer Visium HD set (Tab.~\ref{tab:ablation_loss}). The joint objective achieves the best overall balance, simultaneously reducing errors and improving correlation, indicating that combining fidelity and correlation terms stabilizes training and better aligns predicted spatial patterns with measured gene expression.

\paragraph{Generalization.}

Our network generalizes well to spot sizes different from those used during training. We compare our method's performance with state-of-the-art approaches to evaluate its robustness and adaptability across varying spot sizes, as shown in Tab. \ref{tab:gen}. Our method achieves significantly better performance than the previous competitive method, SGN \cite{sgn}, which is trained to predict gene expression from fixed spots. In contrast, our method reformulates the task as dense prediction, enabling more flexible and accurate gene expression prediction.

\paragraph{Foundation Encoder.}

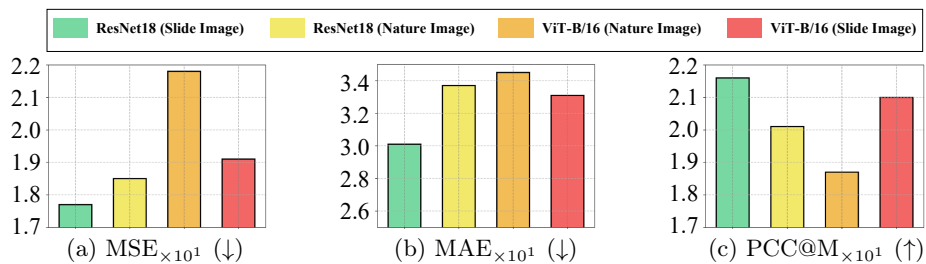
\begin{figure}[t]
    \begin{tikzpicture}
    \tikzset{font=\small}
     \node[anchor=south west,inner sep=0] (image) at (0,0) {\includegraphics[width=\linewidth]{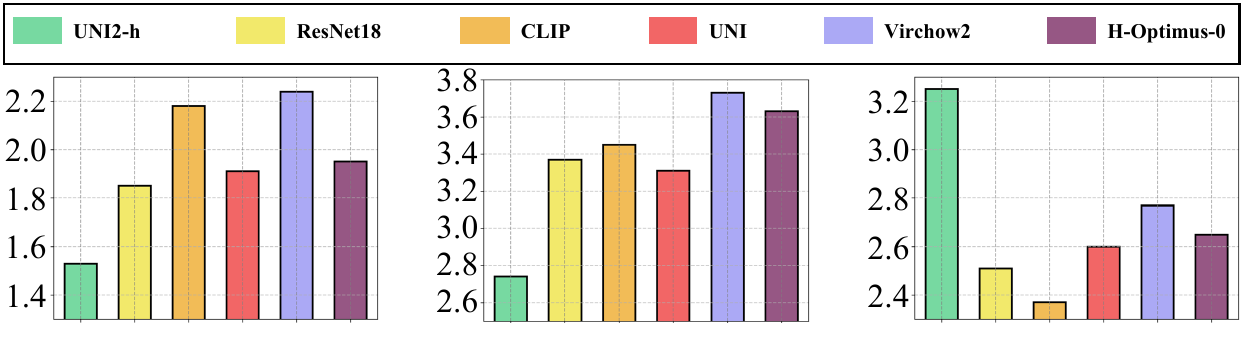}};
        \begin{scope}[x={(image.south east)},y={(image.north west)}]
            \draw (0.16, 0.0) node {(a) $\text{MSE}_{\times 10^1}$ $(\downarrow)$};
            \draw (0.52, 0.0) node {(b) $\text{MAE}_{\times 10^1}$ $(\downarrow)$};
            \draw (0.85 , 0.0) node {(c) $\text{PCC@M}_{\times 10^1}$ $(\uparrow)$};
        \end{scope}
    \end{tikzpicture}
    \caption{Ablation study of the pretrained image encoder \cite{ResNet_pretrain,clip,UNI,zimmermann2024virchow2,hoptimus0}. 
    } 
    \label{fig:ab_encoder}
\end{figure}

We conduct an ablation study on the pretrained image encoder for pyramidal feature extraction, comparing ResNet-18 \cite{ResNet_pretrain} and ViT-L \cite{clip,UNI,zimmermann2024virchow2,hoptimus0}, including pathology foundation models (e.g., Virchow2 \cite{zimmermann2024virchow2}, UNI \cite{UNI}, H-Optimus-0 \cite{hoptimus0}). Results on the breast cancer Visium HD dataset (Fig. \ref{fig:ab_encoder}) show UNI2 \cite{UNI} achieves the best performance.

\paragraph{Spot Size.} We conduct an ablation study on gene expression for spots of different sizes on the breast cancer Visium HD dataset in Tab. \ref{tab:ablation_scale}. The best performance is achieved when all data are utilized.

\section{Limitation}



Despite its strong performance, PixNet still depends on sparse supervision from measured gene expression. Future work will explore unsupervised or self-supervised learning on large-scale histopathology data to achieve more generalizable spatial transcriptomic representations.

\section{Conclusion}
In this paper, we introduce PixNet, for densely predicting the spatial gene expression from histopathology slide images. Our method generates a dense gene expression map from the slide image while applying sparse supervision only to spots with available ground truth gene expression. This approach enables our method to generalize effectively, allowing for accurate gene expression prediction across spots of varying scales and sizes. Extensive experiments demonstrate that our PixNet framework performs competitively compared to state-of-the-art gene expression prediction methods.

{
    \small
    \bibliographystyle{ieeenat_fullname}
    \bibliography{main}
}


\end{document}